%% file: main.tex
\documentclass[runningheads]{llncs}
\usepackage{makeidx}
\usepackage{graphicx}
\usepackage{amsmath,amssymb} % define this before the line numbering.
\usepackage{color}

\usepackage{pgfplots}
\usepackage{filecontents}
\usepackage{hyperref}
\hypersetup{
     colorlinks   = true,
     linkcolor    = blue,
     urlcolor     = black
}

\definecolor{dgreen}{rgb}{0,.7,0}
\definecolor{dyellow}{rgb}{.7,.7,0}
\definecolor{dred}{rgb}{.8,0,0}
\definecolor{dblue}{rgb}{0,0,0.7}
\definecolor{alexey}{rgb}{0.7,0,1}

\include{variables}

\begin{document}
	\pagestyle{headings}
	\mainmatter

	% Replace with your title
	\title{Artistic style transfer for videos}

	\titlerunning{Artistic style transfer for videos}
	\authorrunning{Manuel Ruder, Alexey Dosovitskiy, Thomas Brox}
	\author{Manuel Ruder, Alexey Dosovitskiy, Thomas Brox\thanks{This work was supported by the ERC Starting Grant VideoLearn.}}
	\institute{Department of Computer Science\\University of Freiburg\\ \texttt{\{rudera, dosovits, brox\}@cs.uni-freiburg.de}}

	\maketitle

	\begin{abstract}
	    In the past, manually re-drawing an image in a certain artistic style required a professional artist and a long time.
	    Doing this for a video sequence single-handed was beyond imagination.
	    Nowadays computers provide new possibilities.
	    We present an approach that transfers the style from one image (for example, a painting) to a whole video sequence.
	    We make use of recent advances in style transfer in still images and propose new initializations and loss functions applicable to videos.
	    This allows us to generate consistent and stable stylized video sequences, even in cases with large motion and strong occlusion.
	    We show that the proposed method clearly outperforms simpler baselines both qualitatively and quantitatively.
	\end{abstract}	

	\section{Introduction}
	\label{sec:introduction}
	
	There have recently been a lot of interesting contributions to the issue of style transfer using deep neural networks. Gatys et al.~\cite{DBLP:journals/corr/GatysEB15a} proposed a novel approach using neural networks to capture the  style of artistic images and transfer it to real world photographs. Their approach uses high-level feature representations of the images from hidden layers of the VGG convolutional  network~\cite{DBLP:journals/corr/SimonyanZ14a} to separate and reassemble content and style. This is done by formulating an optimization problem that, starting with white noise, searches for a new image showing similar neural activations as the \textit{content image} and similar feature correlations (expressed by a Gram matrix) as the \textit{style image}.
	
\begin{figure}[t]
    \centering
    \includegraphics[width=1\linewidth]{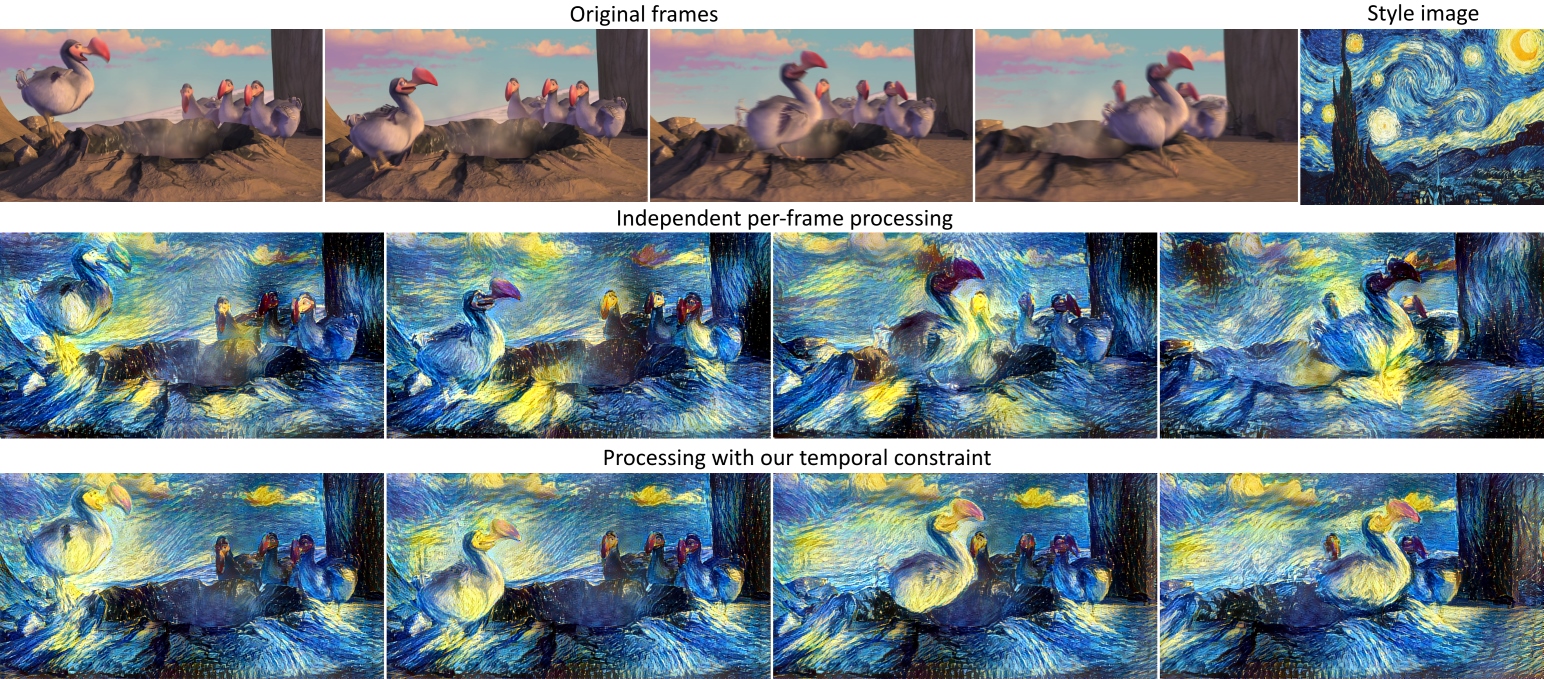}
    \caption{Scene from \emph{Ice Age} (2002) processed in the style of \emph{The Starry Night}. 
    Comparing independent per-frame processing to our time consistent approach, the latter is clearly preferable. Best observed in the supplemental video, see section~\ref{sec:video}.}
    \label{fig:ImageTeaser}
    \end{figure}

	The present paper builds upon the approach from Gatys et al.~\cite{DBLP:journals/corr/GatysEB15a} and extends style transfer to video sequences. Given an artistic image, we transfer its particular style of painting to the entire video. Processing each frame of the video independently leads to flickering and false discontinuities, since the solution of the style transfer task is not stable.
	To regularize the transfer and to preserve smooth transition between individual frames of the video, we introduce a temporal constraint that penalizes deviations between two frames. The temporal constraint takes the optical flow from the original video into account: instead of penalizing the deviations from the previous frame, we penalize deviation along the point trajectories. Disoccluded regions as well as motion boundaries are excluded from the penalizer. This allows the process to rebuild disoccluded regions and distorted motion boundaries while preserving the appearance of the rest of the image, see Fig.~\ref{fig:ImageTeaser}.

	In addition, we present two extensions of our approach. 
	The first one aims on improving the consistency over larger periods of time. When a region that is occluded in some frame and disoccluded later gets rebuilt during the process, most likely this region will have a different appearance than before the occlusion. 
	To solve this, we make use of long term motion estimates.
	This allows us to enforce consistency of the synthesized frames before and after the occlusion.
	
	Secondly, the style transfer tends to create artifacts at the image boundaries. For static images, these artifacts are hardly visible, yet for videos with strong camera motion they move towards the center of the image and get amplified. 
	We developed a multi-pass algorithm, which processes the video in alternating directions using both forward and backward flow.
	This results in a more coherent video.
	
	We quantitatively evaluated our approach in combination with  different optical flow algorithms on the Sintel benchmark.
	Additionally we show qualitative results on several movie shots.
	We were able to successfully eliminate most of the temporal artifacts and can create smooth and coherent stylized videos.
	
	\section{Related work}
	\label{sec:introduction}
	\subsubsection{Style transfer using deep networks:} Gatys et al. \cite{DBLP:journals/corr/GatysEB15a} showed remarkable results by using the VGG-19 deep neural network for style transfer. Their approach was taken up by various follow-up papers that, among other things, proposed different ways to represent the style within the neural network. Li et al. \cite{DBLP:journals/corr/LiW16} suggested an approach to preserve local patterns of the style image. Instead of using a global representation of the style, computed as Gram matrix, they used patches of the neural activation from the style image. Nikulin et al. \cite{DBLP:journals/corr/NikulinN16} tried the style transfer algorithm by Gatys et al. on other nets than VGG and proposed several variations in the way the style of the image is represented to archive different goals like illumination or season transfer.
	However, we are not aware of any work that applies this kind of style transfer to videos.
	
	\textbf{Painted animations:} One common approach to create video sequences with an artistic style is to generate artificial brush strokes to repaint the scene. Different artistic styles are gained by modifying various parameters of these brush strokes, like thickness, or by using different brush placement methods. To achieve temporal consistency Litwinowicz \cite{Litwinowicz:1997:PIV:258734.258893} was one of the first who used optical flow. In his approach, brush strokes were generated for the first frame and then moved along the flow field. Later, this approach was refined. Hays et al. \cite{Hays:2004:IVB:987657.987676} proposed new stylistic parameters for the brush strokes to mimic different artistic styles. O'Donovan et al. \cite{odonovan:2012} formulated an energy optimization problem for an optimal placement and shape of the brush strokes and also integrated a temporal constraint into the optimization problem by penalizing changes in shape and width of the brush strokes compared to the previous frame. 
	These approaches are similar in spirit to what we are doing, but they are only capable of applying a restricted class of artistic styles.

   \section{Style transfer in still images}
   \label{sec:methods}
   
   In this section, we briefly review the style transfer approach introduced by Gatys et al.~\cite{DBLP:journals/corr/GatysEB15a}. The aim is to generate a stylized image $\gen$ showing the content of an image $\img$ in the style of an image $\styl$.
   Gatys et al. formulated an energy minimization problem consisting of a \textit{content loss} and a \textit{style loss}.
   The key idea is that features extracted by a convolutional network carry information about the content of the image, while the correlations of these features encode the style. 
    
   We denote by $\net^l (\cdot)$ the function implemented by the part of the convolutional network from input up to the layer $l$.
    The feature maps extracted by the network from the original image $\img$, the style image $\styl$ and the stylized image $\gen$ we denote by $\vec{P}^l = \net^l (\img)$, $\vec{S}^l = \net^l(\styl)$ and $\vec{F}^l = \net^l(\gen)$ respectively.
    The dimensionality of these feature maps we denote by $N_l \times M_l$, where $N_l$ is the number of filters (channels) in the layer, and $M_l$ is the spatial dimensionality of the feature map, that is, the product of its width and height.
    
    The content loss, denoted as $\loss_{content}$, is simply the mean squared error between 
    $\vec{P}^l \in \mathbb{R}^{N_l\times M_l}$ 
    and $\vec{F}^l \in \mathbb{R}^{N_l\times M_l}$. 
    This loss need not be restricted to only one layer. Let $L_{content}$  be the set of layers to be used for content representation, then we have:
    \begin{equation}
    \loss_{content}\bigl(\vec{p}, \vec{x}\bigr) = \sum_{l \in L_{content}}  \frac{1}{N_l M_l} \sum_{i,j} \bigr( F^l_{ij} - P^l_{ij} \bigl)^2.
    \end{equation}
    
    The style loss is also a mean squared error, but between the correlations of the filter responses expressed by their Gram matrices $A^l \in \mathbb{R}^{N_l\times N_l}$ for the style image $\styl$ and $G^l \in \mathbb{R}^{N_l\times N_l}$ for the stylized image $\gen$.
    These are computed as
    $A^l_{ij} = \sum\limits_{k=1}^{M_l} S^l_{ik} S^l_{jk}$ and 
    $G^l_{ij} = \sum\limits_{k=1}^{M_l} F^l_{ik} F^l_{jk}\,$.
    As above, let  $L_{style}$ be the set of layers we use to represent the style, then the style loss is given by:
    \begin{equation}
    \mathcal{L}_{style}\bigl(\vec{a}, \vec{x}\bigr) = \sum_{l \in L_{style}} \frac{1}{N_l^2 M_l^2} \sum_{i,j} \bigr( G^l_{ij} - A^l_{ij} \bigl)^2 
    \end{equation}
    Overall, the loss function is given by
    \begin{equation}
	\mathcal{L}_{singleimage}\bigl(\vec{p}, \vec{a}, \vec{x}\bigr) =
	\alpha \mathcal{L}_{content}\bigl(\vec{p}, \vec{x}\bigr) \ + \
	\beta \mathcal{L}_{style}\bigl(\vec{a}, \vec{x}\bigr),
	\end{equation}
	with weighting factors $\alpha$ and $\beta$ governing the importance of the two components.
	
    The stylized image is computed by minimizing this energy with respect to $\gen$ using gradient-based optimization.
    Typically it is initialized with random Gaussian noise.
    However, the loss function is non-convex, therefore the optimization is prone to falling into local minima. 
    This makes the initialization of the stylized image important, especially when applying the method to frames of a video.

    \section{Style transfer in videos}
    
    We use the following notation: $\vec{p}^{(i)}$ is the $i^{th}$ frame of the original video, $\vec{a}$ is the style image and $\vec{x}^{(i)}$ are the stylized frames to be generated. Furthermore, we denote by $\vec{x'}^{(i)}$ the initialization of the style optimization algorithm at frame $i$. 
   By $x_j$ we denote the $j^{th}$ component of a vector $\vec{x}$.
    
    \subsection{Short-term consistency by initialization}
    
    When the style transfer for consecutive frames is initialized by independent Gaussian noise, two frames of a video converge to very different local minima, resulting in a strong flickering.
    The most basic way to yield temporal consistency is to initialize the optimization for the frame $i+1$ with the stylized frame $i$. Areas that have not changed between the two frames are then initialized with the desired appearance, while the rest of the image has to be rebuilt through the optimization process.
    
    If there is motion in the scene, this simple approach does not perform well, since moving objects are initialized incorrectly.
    Thus, we take the optical flow into account and initialize the optimization for the frame $i+1$ with the previous stylized frame warped: $\gen'^{(i+1)} = \omega_i^{i+1}\bigl(\gen^{(i)}\bigr)$. 
    Here $\omega_{i}^{i+1}$ denotes the function that warps a given image using the optical flow field that was estimated between image $\imgn{i}$ and $\imgn{i+1}$.
    Clearly, the first frame of the stylized video still has to be initialized randomly.
    
    We experimented with two state-of-the-art optical flow estimation algorithms: DeepFlow \cite{weinzaepfel:hal-00873592} and EpicFlow \cite{revaud:hal-01142656}.
    Both are based on Deep Matching~\cite{weinzaepfel:hal-00873592}: DeepFlow combines it with a variational approach, while EpicFlow relies on edge-preserving sparse-to-dense interpolation. 
    
    \subsection{Temporal consistency loss}
    To enforce stronger consistency between adjacent frames we additionally introduce an explicit consistency penalty to the loss function.
    This requires detection of disoccluded regions and motion boundaries.
    To detect disocclusions, we perform a forward-backward consistency check of the optical flow \cite{Bro10e}.
    Let $\flow = (u, v)$ be the optical flow in forward direction and $\hat{\flow} =(\hat{u}, \hat{v})$ the flow in backward direction.
    Denote by $\widetilde{\flow}$ the forward flow warped to the second image:
    \begin{equation}
    \widetilde{\flow} (x,y) = \flow((x,y) + \hat{\flow}(x,y)).
    \end{equation}
   
    In areas without disocclusion, this warped flow should be approximately the opposite of the backward flow. Therefore we mark as disocclusions those areas where the following inequality holds:
    \begin{equation} \label{eq:disoccl}
    |\widetilde{\flow}+ \hat{\flow}|^2 > 0.01 (|\widetilde{\flow}|^2 + |\hat{\flow}|^2) + 0.5
    \end{equation}
    Motion boundaries are detected using the following inequality:
    \begin{equation} \label{eq:motionb}
    |\nabla \vec{\hat{u}}|^2 + |\nabla \vec{\hat{v}}|^2 > 0.01 |\vec{\mathrm{\hat{w}}}|^2  + 0.002
    \end{equation}
    Coefficients in inequalities~\eqref{eq:disoccl} and~\eqref{eq:motionb} are taken from Sundaram et al.~\cite{Bro10e}.
	
	The temporal consistency loss function penalizes deviations from the warped image in regions where the optical flow is consistent and estimated with high confidence:
	\begin{equation}
	\mathcal{L}_{temporal}(\vec{x}, \vec{\omega}, \cert) = \frac{1}{D} \sum_{k=1}^D \certi_k \cdot (x_k - \omega_k)^2\;.
	\end{equation}
    Here $\cert \in [0, 1]^D$ is per-pixel weighting of the loss and $D = W \times H \times C$ is the dimensionality of the image. We define the weights $\cert^{(i-1,i)}$ between frames $i\!-\!1$ and $i$ as follows: $0$ in disoccluded regions (as detected by forward-backward consistency) and at the motion boundaries, and $1$ everywhere else. 
    Potentially weights between $0$ and $1$ could be used to incorporate the certainty of the optical flow prediction.
	The overall loss takes the form:
	\begin{align}
	\loss_{shortterm}\bigl(\imgn{i}, \styl, \genn{i}\bigr) =& \
	\alpha \mathcal{L}_{content}\bigl(\imgn{i}, \genn{i} \bigr) \ + \
	\beta \loss_{style}\bigl(\styl, \genn{i}\bigr) \notag\\
	&+ \ \gamma \loss_{temporal}\bigl(\genn{i},  \omega_{i-1}^i(\genn{i-1}), \cert^{(i-1,i)}\bigr)\;.
	\end{align}
	
	We optimize one frame after another, thus $\genn{i-1}$ refers to the already stylized frame $i\!-\!1$.
	
	Furthermore we experimented with the more robust absolute error instead of squared error for the temporal consistency loss; results are shown in section~\ref{sec:robust_loss}.

	\subsection{Long-term consistency}
	The short-term model has the following limitation: when some areas are occluded in some frame and disoccluded later, these areas will likely change their appearance in the stylized video. This can be counteracted by also making use of long-term motion, i.e. not only penalizing deviations from the previous frame, but also from temporally more distant frames. Let $J$ denote the set of indices each frame should take into account, relative to the frame number. E.g. $J = \{1, 2, 4\}$ means frame $i$ takes frames $i\!-\!1$, $i\!-\!2$ and $i\!-\!4$ into account. Then, the loss function with long-term consistency is given by:
	\begin{align}
	\mathcal{L}_{longterm}\bigl(\vec{p}^{(i)}, \vec{a}, \vec{x}^{(i)}\bigr) =& \
	\alpha \mathcal{L}_{content}\bigl(\vec{p}^{(i)}, \vec{x}^{(i)}\bigr) \ + \
	\beta \mathcal{L}_{style}\bigl(\vec{a}, \vec{x}^{(i)}\bigr) \notag\\
	&+ \ \gamma \sum_{j\in J: i-j \ge 1} \mathcal{L}_{temporal}\bigl(\vec{x}^{(i)},  \omega_{i-j}^i(\vec{x}^{(i-j)}), \certlong^{(i-j,i)}\bigr)
	\end{align}
	It is essential how the weights $\certlong^{(i-j,i)}$ are computed. Let $\cert^{(i-j,i)}$ be the weights for the flow between image $i\!-\!j$ and $i$, as defined for the short-term model. 
	The long-term weights $\certlong^{(i-j,i)}$ are computed as follows:
	\begin{equation}
	\certlong^{(i-j,i)} = \max \bigl( \cert^{(i-j,i)} - \sum_{k\in J: i-k > i-j} \cert^{(i-k,i)},\; \vec{0} \bigr) \; ,
    \end{equation}
    where $\max$ is taken element-wise.
    This means, we first apply the usual short-term constraint. For pixels in disoccluded regions we look into the past until we find a frame in which these have consistent correspondences.
    An advantage over simply using $ \cert^{(i-j,i)}$ is that each pixel is connected only to the closest possible frame from the past. 
    Since the optical flow computed over more frames is more erroneous than over fewer frames, this results in nicer videos. An empirical comparison of $\cert^{(i-j,i)}$ and $\certlong^{(i-j,i)}$ is shown in the supplementary video (see section~\ref{sec:video}).

    \subsection{Multi-pass algorithm}
    
    We found that the output image tends to have less contrast and is less diverse near image boundaries than in other areas of the image. For mostly static videos this effect is hardly visible. However, in cases of strong camera motion the areas from image boundaries move towards the center of the image, which leads to a lower image quality over time when combined with our temporal constraint. Therefore, we developed a multi-pass algorithm which processes the whole sequence in multiple passes and alternating directions. The basic idea is that we progressively propagate intermediate results in both forward and backward direction, avoiding an one-way information flow from the image boundaries to the center only.
    
    Every pass consists of a relatively low number of iterations without full convergence. At the beginning, we process every frame independently based on a random initialization. After that, we blend frames with non-disoccluded parts of previous frames warped according to the optical flow, then run the optimization algorithm for some iterations initialized with this blend. The direction in which the sequence is processed is alternated in every pass. We repeat this blending and optimization to convergence.
    
    Formally, let $\gen'^{(i)(j)}$ be the initialization of frame $i$ in pass $j$ and $\gen^{(i)(j)}$ the corresponding output after some iterations of the optimization algorithm.
    When processed in forward direction, the initialization of frame $i$ is created as follows:
    \begin{equation}
    \vec{x'}^{(i)(j)} = 
     \begin{cases}
     \vec{x}^{(i)(j-1)} &\text{if } i=1,\\
     \delta \cert^{(i-1,i)} \circ \omega_{i-1}^{i}\bigl(\vec{x}^{(i-1)(j)}\bigr)
     + (\overline{\delta} \vec{1} + \delta \overline{\cert}^{(i-1,i)}) \circ \vec{x}^{(i)(j-1)}
     &\text{else}.
     \end{cases}
    \end{equation}
    Here $\circ$ denotes element-wise vector multiplication, $\delta$ and $\overline{\delta} = 1\!-\!\delta$ are the blend factors, $\vec{1}$ is a vector of all ones, and $\overline{\cert} = \vec{1} - \cert$.
    
    Analogously, the initialization for a backward direction pass is:
    \begin{equation}
    \vec{x'}^{(i)(j)} = 
     \begin{cases}
     \vec{x}^{(i)(j-1)} &\text{if } i=N_{\text{frames}}\\
     \delta \cert^{(i+1,i)} \circ \omega_{i+1}^{i}\bigl(\vec{x}^{(i+1)(j)}\bigr)
     + (\overline{\delta} \vec{1} + \delta \overline{\cert}^{(i+1,i)}) \circ \vec{x}^{(i)(j-1)}
     &\text{else}
     \end{cases}
    \end{equation}
    
    The multi-pass algorithm can be combined with the temporal consistency loss described above. We achieved good results when we disabled the temporal consistency loss in several initial passes and enabled it in later passes only after the images had stabilized.
    
    \begin{figure}[!ht]
    \centering
    \vspace{-0.5cm}
    \includegraphics[width=0.75\linewidth]{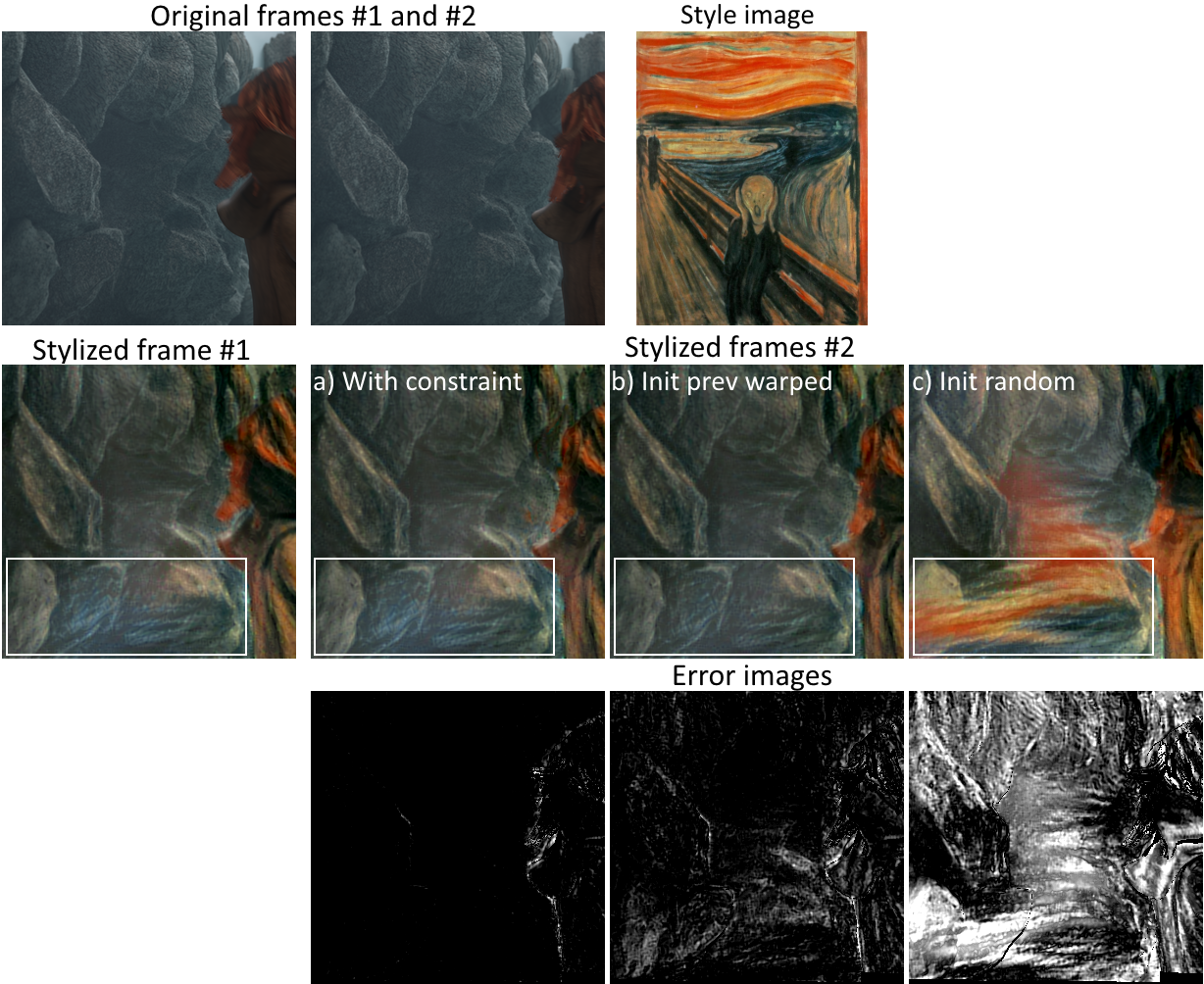}
    \caption{Close-up of a scene from Sintel, combined with \emph{The Scream} painting. \textbf{a)} With temporal constraint \textbf{b)} Initialized with previous image warped, but without the constraint \textbf{c)} Initialized randomly. The marked regions show most visible differences. \emph{Error images} show the contrast-enhanced absolute difference between frame \#1 and frame \#2 warped back using ground truth optical flow, as used in our evaluation. The effect of the temporal constraint is very clear in the error images and in the corresponding video.}
    \vspace{-0.5cm}
    \label{fig:sintel_qualitative}
    \end{figure}
   
    \section{Experiments}
    
    In this section, we briefly describe implementation details and present experimental results produced with different versions of our algorithm.
    While we did our best to make the paper self-contained, it is not possible to demonstrate effects like video flickering in still images.
	We therefore advise the readers to watch the supplementary video, which is available at \url{https://youtu.be/vQk_Sfl7kSc}.
    
    \subsection{Implementation details}
    
    Our implementation\footnote{GitHub: \url{https://github.com/manuelruder/artistic-videos}} is based on the Torch~\cite{Collobert_NIPSWORKSHOP_2011} implementation called \emph{neural-style}\footnote{GitHub: \url{https://github.com/jcjohnson/neural-style}}.
    We used the following layers of the VGG-19 network~\cite{DBLP:journals/corr/SimonyanZ14a} for computing the losses: \emph{relu4\_2} for the content and \emph{relu1\_1},\emph{relu2\_1},\emph{relu3\_1},\emph{relu4\_1},\emph{relu5\_1} for the style. The energy function was minimized using L-BFGS. 
    For precise evaluation we incorporated the following strict stopping criterion: the optimization was considered converged if the loss did not change by more than $0.01\%$ during 50 iterations.
    This typically resulted in roughly $2000$ to $3000$ iterations for the first frame and roughly $400$ to $800$ iterations for subsequent frames when optimizing with our temporal constraint, depending on the amount of motion and the complexity of the style image.
    Using a convergence threshold of $0.1\%$  cuts the number of iterations and the running time in half, and we found it still produces reasonable results in most cases. However, we used the stronger criterion in our experiments for the sake of accuracy.

    For videos of resolution $350 \times 450$ we used weights $\alpha = 1$ and $\beta = 20$ for the content and style losses, respectively (default values from \emph{neural-style}),
    and weight $\gamma = 200$ for the temporal losses. 
    However, the weights should be adjusted if the video resolution is different.
    We provide the details in section~\ref{sec:weightings}.
    
    For our multi-pass algorithm, we used $100$ iterations per pass and set $\delta = 0.5$, but we needed at least $10$ passes for good results, so this algorithm needs more computation time than our previous approaches.
    
    We used DeepMatching, DeepFlow and EpicFlow implementations provided by the authors of these methods. 
    We used the "improved-settings" flag in DeepMatching 1.0.1 and the default settings for DeepFlow 1.0.1 and EpicFlow 1.00.
    
    \subsubsection{Runtime} For the relaxed convergence threshold of $0.1\%$ with random initialization the optimization process needed on average roughly eight to ten minutes per frame at a resolution of $1024 \times 436$ on an Nvidia Titan X GPU.
    When initialized with the warped previous frame and combined with our temporal loss, the optimization converges 2 to 3 times faster, three minutes on average.
    Optical flow computation runs on a CPU and takes roughly $3$ minutes per frame pair (forward and backward flow together), therefore it can be performed in parallel with the style transfer.
    Hence, our modified algorithm is roughly $3$ times faster than naive per-frame processing, while providing temporally consistent output videos.

    \subsection{Short-term consistency}

    We evaluated our short-term temporal loss  on $5$ diverse scenes from the  MPI Sintel Dataset~\cite{Butler:ECCV:2012}, with 20 to 50 frames of resolution $1024 \times 436$ pixels per scene, and $6$ famous paintings (shown in section~\ref{sec:paintings}) as style images. The Sintel dataset provides ground truth optical flow and ground truth occlusion areas, which allows a quantitative study. We warped each stylized frame $i$ back with the ground truth flow and computed the difference with the stylized frame $i-1$ in non-disoccluded regions.
    We use the mean square of this difference (that is, the mean squared error) as a quantitative performance measure.
    
    On this benchmark we compared several approaches: our short-term consistency loss with DeepFlow and EpicFlow, as well as three different initializations without the temporal loss: random noise, the previous stylized frame and the previous stylized frame warped with DeepFlow. We set $\alpha = 1$, $\beta = 100$, $\gamma = 400$. 
    
    A qualitative comparison is shown in Fig.~\ref{fig:sintel_qualitative}. Quantitative results are in Table~\ref{table:benchmark}. The most straightforward approach, processing every frame independently, performed roughly an order of magnitude worse than our more sophisticated methods. In most cases, the temporal penalty significantly improved the results.
    The \emph{ambush} scenes are exceptions, since they contain very large motion and the erroneous optical flow impairs the temporal constraint.
    Interestingly, on average DeepFlow performed slightly better than EpicFlow in our experiments, even through EpicFlow outperforms DeepFlow on the Sintel optical flow benchmark.

    \setlength{\tabcolsep}{4pt}
	\begin{table}
	    \vspace{-0.3cm}
		\centering
		\caption{Short-term consistency benchmark results. Mean squared error of different methods on $5$ video sequences, averaged over $6$ styles, is shown. Pixel values in images were between $0$ and $1$.}
		\label{table:benchmark}
		\begin{tabular}{l|lllll|l}
			\hline\noalign{\smallskip}
			 & alley\_2 & ambush\_5 & ambush\_6 & bandage\_2 & market\_6\\
			\noalign{\smallskip}
			\hline
			\noalign{\smallskip}
			\hline
			DeepFlow          & \textbf{0.00061} & \textbf{0.0062} & \textbf{0.012} & 0.00084 & 0.0035 \\
			EpicFlow          & 0.00073 & 0.0068 & 0.014 & \textbf{0.00080} & \textbf{0.0032} \\
			Init prev warped  & 0.0016 & 0.0063 & \textbf{0.012} & 0.0015 & 0.0049 \\
			Init prev         & 0.010 & 0.018 & 0.028 & 0.0041 & 0.014 \\
			Init random       & 0.019 & 0.027 & 0.037 & 0.018 & 0.023 \\
			\hline
		\end{tabular}
		\vspace{-1.0cm}
	\end{table}
	\setlength{\tabcolsep}{1.4pt}
    
    \subsection{Long-term consistency and multi-pass algorithm}
    The short-term consistency benchmark presented above cannot evaluate the long-term consistency of videos (since we do not have long-term ground truth flow available) and their visual quality (this can only be judged by humans). 
    We therefore excluded the long-term penalty and the multi-pass approach from the quantitative comparison and present only qualitative results. Please see the supplementary video for more results.
    
\begin{figure}
    \centering
    \vspace{-0.5cm}
    \includegraphics[width=0.85\linewidth]{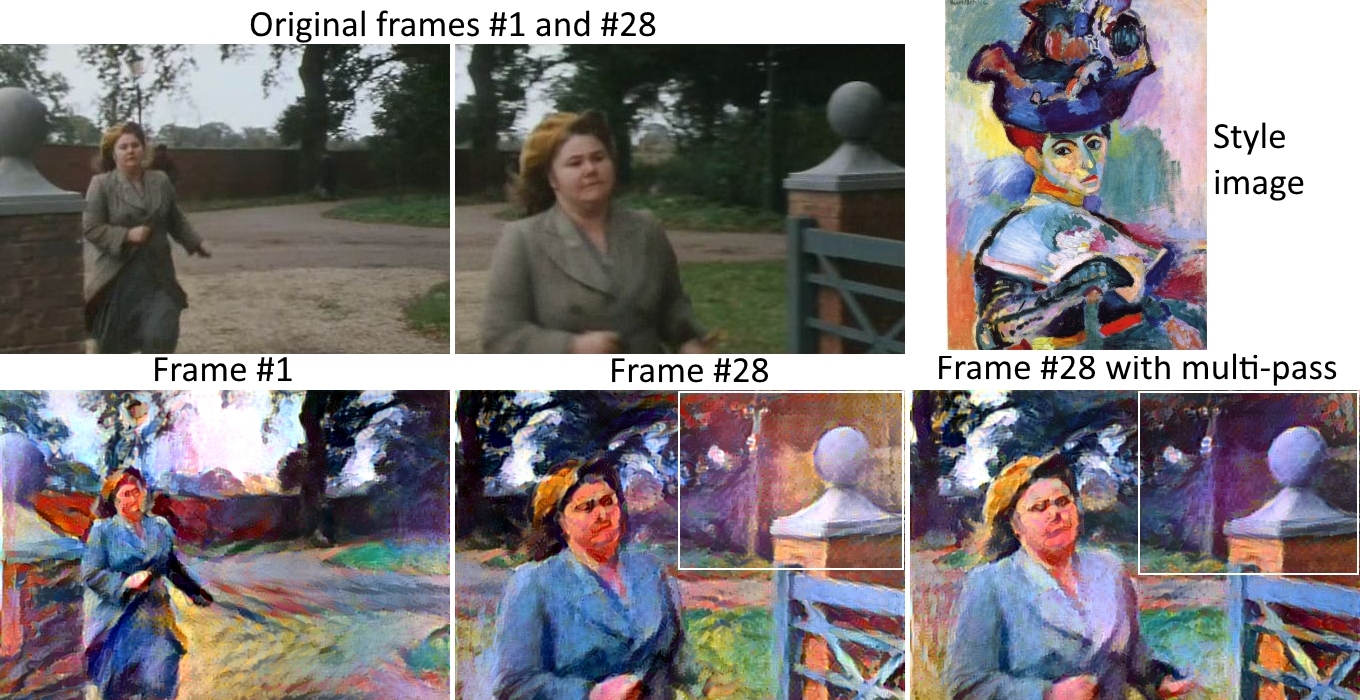}
    \caption{The multi-pass algorithm applied to a scene from Miss Marple. With the default method, the image becomes notably brighter and loses contrast, while the multi-pass algorithm yields a more consistent image quality over time. 
    Corresponding video is linked in section~\ref{sec:video}.
    }
    %\vspace{-1.5cm}
    \label{fig:ImageMultiPass}
    \end{figure}

    Fig.~\ref{fig:ImageLongTerm} shows a scene from Miss Marple where a person walks through the scene. Without our long-term consistency model, the background looks very different after the person passes by. The long-term consistency model keeps the background unchanged. Fig.~\ref{fig:ImageMultiPass} shows another scene from Miss Marple with fast camera motion. The multi-pass algorithm avoids the artifacts introduced by the basic algorithm.
    
    \section{Conclusion}
    We presented a set of techniques for style transfer in videos: suitable initialization, a loss function that enforces short-term temporal consistency of the stylized video, a loss function for long-term consistency, and a multi-pass approach.
    As a consequence, we can produce stable and visually appealing stylized videos even in the presence of fast motion and strong occlusion.

    \begin{figure}
    \centering
    \vspace{-0.4cm}
    \includegraphics[width=0.85\linewidth]{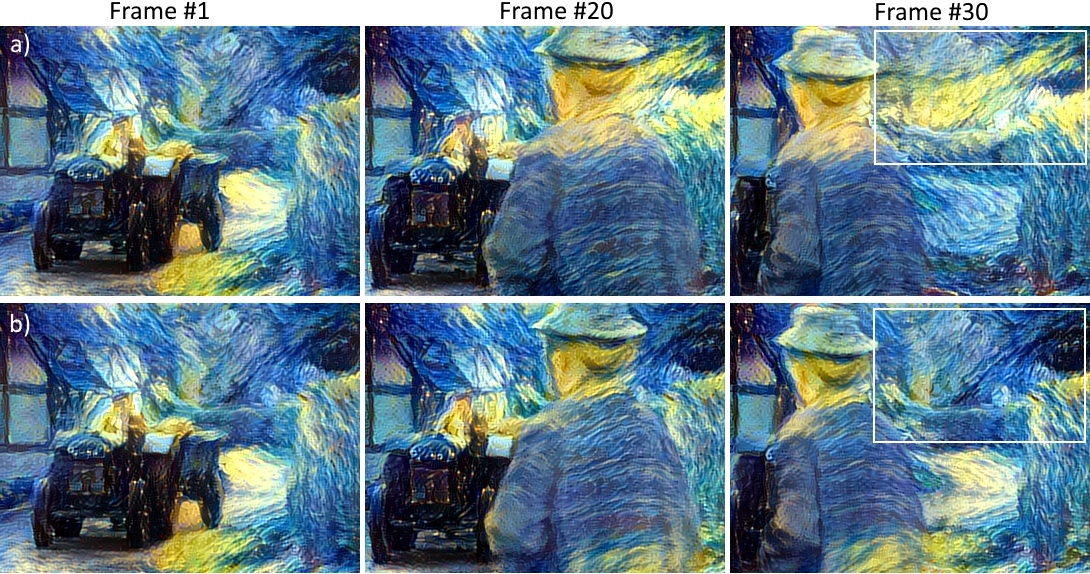}
    \caption{Scene from Miss Marple, combined with The Starry Night painting. \textbf{a)} Short-term consistency only. \textbf{b)} Long-term consistency with $J = \{1, 10, 20, 40\}$.
    Corresponding video is linked in section~\ref{sec:video}.}
    \label{fig:ImageLongTerm}
    \vspace{-0.4cm}
    \end{figure}
    
	\bibliographystyle{splncs03}
	\bibliography{references}
	
	\newpage
	
	\section*{Supplementary material}
	\input{supplementary}

\end{document}

%% file: variables.tex
    \newcommand{\loss}{\mathcal{L}}
    \newcommand{\img}{\vec{p}}
    \newcommand{\imgn}[1]{\img^{(#1)}}
    \newcommand{\gen}{\vec{x}}
    \newcommand{\genn}[1]{\gen^{(#1)}}
    \newcommand{\styl}{\vec{a}}
    \newcommand{\net}{\Phi}
    \newcommand{\flow}{\vec{\mathrm{w}}}
    \newcommand{\cert}{\vec{c}}
    \newcommand{\certi}{c}
    \newcommand{\certlong}{\vec{c}_{long}}

%% file: supplementary.tex
		% Maybe related work:
	% * Video-Based Running Water Animation in Chinese Painting Style
	%   http://cg.cs.tsinghua.edu.cn/papers/tr080801.pdf
	% * AniPaint: Interactive Painterly Animation from Video
	%   http://www.dgp.toronto.edu/~donovan/anipaint/anipaint.pdf
	% * "Processing Images and Video for An Impressionist Effect", Litwinowicz 1997
	%   Not public, but can be found somewhere on the internet
	% * Image and Video Based Painterly Animation
	%   http://www.cc.gatech.edu/~hays/papers/IVBPA_Final.pdf
	
	\section{Additional details of experimental setup}
	
	\subsection{Style images}
	\label{sec:paintings}
	
\begin{figure}
    \centering
    \includegraphics[width=0.8\linewidth]{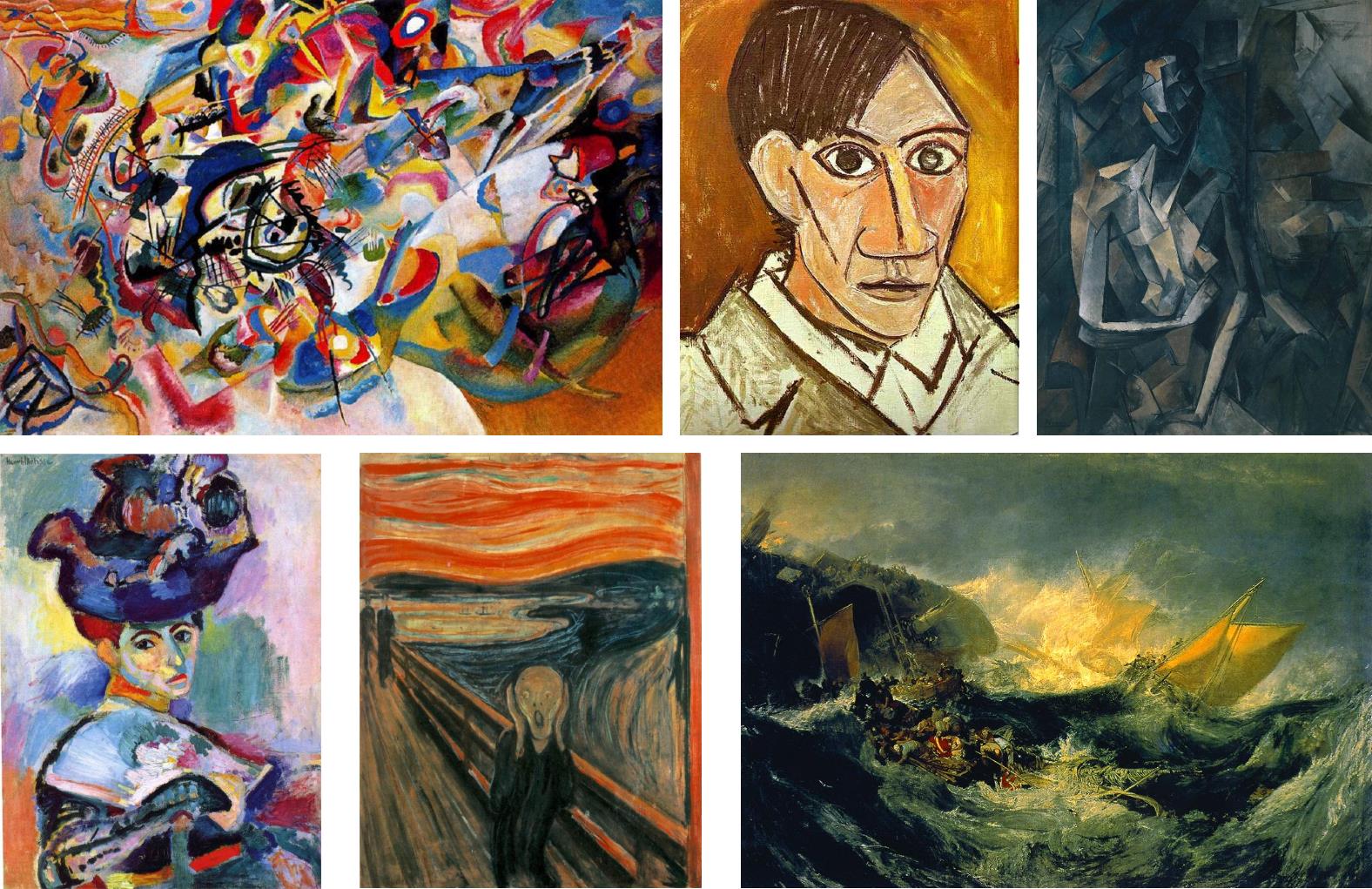}
    \caption{Styles used for experiments on Sintel. Left to right, top to bottom: "Composition VII" by Wassily Kandinsky (1913), Self-Portrait by Pablo Picasso (1907), "Seated female nude" by Pablo Picasso (1910), "Woman with a Hat" by Henri Matisse (1905), "The Scream" by Edvard Munch (1893), "Shipwreck" by William Turner (1805).}
    \label{fig:styles_benchmark}
    \end{figure}

	Style images we used for benchmark experiments on Sintel are shown in Figure~\ref{fig:styles_benchmark}.
	
    \subsection{Weighting of the loss components}
    \label{sec:weightings}
    
    As mentioned in the main paper, for best results the weights $\alpha$, $\beta$ and $\gamma$ of different components of the loss function have to be adjusted depending on the resolution of the video.
    The settings we used for different resolutions are shown in Table~\ref{tbl:loss_weights}.
    
	\begin{table}
		\centering
		\caption{Weights of the loss function components for different input resolutions.}
		\label{tbl:loss_weights}
		\label{table:benchmark}
		\begin{tabular}{l|ccc}
		     & $\ 350 \times 450\ $ &\ $768 \times 432\ $ & $\ 1024 \times 436\ $ \\
			\hline
			$\alpha$ (content) & 1 & 1 & 1 \\
			$\beta$ (style) & 20 & 40 & 100 \\
			$\gamma$ (temporal) & 200 & 200 & 400
		\end{tabular}
	\end{table}
	
	\section{Additional experiments}
	\label{sec:robust_loss}
	
	\subsection{Supplementary video}
    \label{sec:video}
	A supplementary video, available at \url{https://youtu.be/vQk_Sfl7kSc}, shows moving sequences corresponding to figures from this paper, plus a number of additional results:
	\begin{itemize}
	\item Results of the basic algorithm on different sequences from Sintel with different styles
	\item Additional comparison of the basic and the multi-pass algorithm
	\item Additional comparison of the basic and the long-term algorithm
	\item Comparison of "naive" ($\cert$) and "advanced" ($\certlong$) weighting schemes for long-term consistency
	
	Another video, showing results of the algorithm on a number of diverse videos with different style images is avalable at \url{https://youtu.be/Khuj4ASldmU}.
	\end{itemize}
	
	\subsection{Robust loss function for temporal consistency}
	
\begin{figure}
    \centering
    \includegraphics[width=0.9\linewidth]{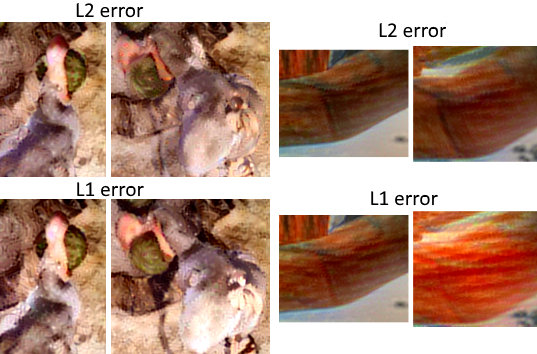}
    \caption{\textbf{Left}: Scene from Ice Age (2002) where an absolute error function works better, because the movement of the bird wasn't captured correctly by the optical flow. \textbf{Right}: Extreme case from Sintel movie where a squared error is far superior.}
    \label{fig:Image3}
    \end{figure}

	We tried using the more robust absolute error instead of squared error for the temporal consistency loss. The weight for the temporal consistency was doubled in this case.
	Results are shown in Figure~\ref{fig:Image3}.
	While in some cases (left example in the figure) absolute error leads to slightly improved results, in other cases (right example in the figure) it causes large fluctuations.
	We therefore stick with mean squared error in all our experiments.
	
    \subsection{Effect of errors in optical flow estimation}
    
\begin{figure}
    \centering
    \includegraphics[width=1\linewidth]{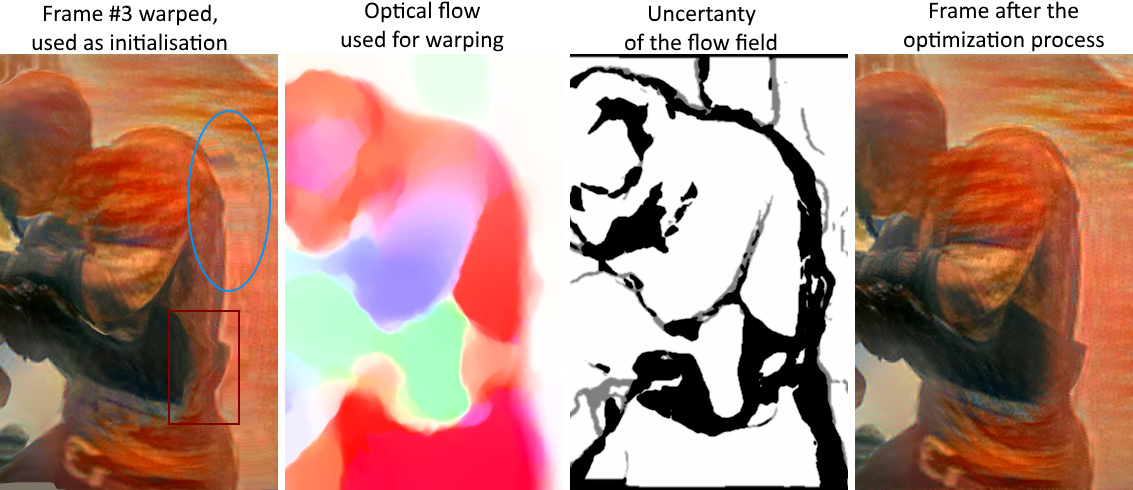}
    \caption{Scene from the Sintel video showing how the algorithm deals with optical flow errors (red rectangle) and disocclusions (blue circle). Both artifacts are somehow repaired in the optimization process due to the exclusion of uncertain areas from our temporal constrain.
    Still, optical flow errors lead to imperfect results. The third image shows the uncertainty of the flow filed in black and motion boundaries in gray.}
    \label{fig:Image2}
    \end{figure}

    The quality of results produced by our algorithm strongly depends on the quality of optical flow estimation.
    This is illustrated Figure~\ref{fig:Image2}. 
    When the optical flow is correct (top right region of the image), the method manages to repair the artifacts introduced by warping in the disoccluded region.
    However, erroneous optical flow (tip of the sword in the bottom right) leads to degraded performance. Optimization process partially compensates the errors (sword edges get sharp), but cannot fully recover.

